\documentclass[runningheads]{llncs}
\usepackage[T1]{fontenc}
\usepackage{graphicx,verbatim}
\usepackage{xcolor}
\usepackage{multirow} 
\usepackage{booktabs}
\usepackage{hyperref}

\begin{document}
\title{Semantic Scene Graph for Ultrasound Image Explanation and Scanning Guidance
}
%

\author{Xuesong Li\inst{1, 2} \and Dianye Huang\inst{1, 2} \and Yameng Zhang\inst{1, 3} \and \\ Nassir Navab\inst{1} \and Zhongliang Jiang\inst{1}}  
\authorrunning{Xuesong Li et al.}
\titlerunning{Semantic Scene Graph for Ultrasound Image}
\institute{Computer Aided Medical Procedures (CAMP),\\ TU Munich, Germany \\
\email{zl.jiang@tum.com} \\
\and Munich Center for Machine Learning (MCML), Munich, Germany 
\and The Chinese University of Hong Kong, Hong Kong, China}

\maketitle              
\begin{abstract}
Understanding medical ultrasound imaging remains a long-standing challenge due to significant visual variability caused by differences in imaging and acquisition parameters. Recent advancements in large language models (LLMs) have been used to automatically generate terminology-rich summaries orientated to clinicians with sufficient physiological knowledge. Nevertheless, the increasing demand for improved ultrasound interpretability and basic scanning guidance among non-expert users, e.g., in point-of-care settings, has not yet been explored. 
In this study, we first introduce the scene graph (SG) for ultrasound images to explain image content to ordinary and provide guidance for ultrasound scanning. The ultrasound SG is first computed using a transformer-based one-stage method, eliminating the need for explicit object detection. To generate a graspable image explanation for ordinary, the user query is then used to further refine the abstract SG representation through LLMs. Additionally, the predicted SG is explored for its potential in guiding ultrasound scanning toward missing anatomies within the current imaging view, assisting ordinary users in achieving more standardized and complete anatomical exploration. The effectiveness of this SG-based image explanation and scanning guidance has been validated on images from the left and right neck regions, including the carotid and thyroid, across five volunteers. The results demonstrate the potential of the method to maximally democratize ultrasound by enhancing its interpretability and usability for ordinaries. Project page: \url{https://noseefood.github.io/us-scene-graph/}











\end{abstract}
%
%
%


\section{Introduction}
%
\par
Medical ultrasound (US) is widely used in modern clinical practice for examining internal organs such as the carotid, thyroid, and liver. With its accessibility and portability, US imaging has the potential for widespread deployment, making it more universally available. However, interpreting US images requires substantial experience due to significant visual variability stemming from differences in imaging and acquisition parameters. Unlike CT and MRI, US interpretation is less intuitive due to its limited field of view and lack of 3D structural information. Therefore, effective US image explanation and scanning guidance are essential for broader adoption, particularly in point-of-care ultrasound (POCUS) settings~\cite{ockenden2024role,andersen2019point,jiang2023robotic,bi2024machine} and for self-learning of anatomy and physiology among non-experts.


\par
Drawing inspiration from scene graph (SG) technology in classical computer vision~\cite{dhamo2020semantic,yang2018graph,im2024egtr} and emerging surgical data science~\cite{maier2017surgical,ozsoy20224d,murali2023latent,yuan2024advancing,islam2020learning}, this approach effectively summarizes key objects and their relationships within images. Therefore, an intuitive image explanation can be generated by leveraging the relations defined in the predicted SG. Unlike recent efforts focused on comprehensive report generation using large language models (LLMs)~\cite{huh2023breast,li2024ultrasound}, computing a conceptualized SG representation for individual images offers greater flexibility. A comprehensive US report with medical terminology may be demanded by clinicians but can be less intuitive for non-expert users. A recent effort pioneering introducing SG to describe the objects and their relations in CT image has been reported in~\cite{sanner2024voxel}. In contrast to a comprehensive summary, an SG representation provides a highly conceptualized summary, emphasizing only key information. This intermediate representation can be seamlessly adapted for various downstream tasks, such as US image summarization or probe motion guidance for medical student training, by further integrating the full-level prior anatomical knowledge. 

\par
Ultrasound is a 2D cross-sectional image, unlike natural images that feature a distinct foreground and background. This characteristic, along with the relatively stable anatomical structure, simplifies object relationship extraction compared to natural images. Due to the low-contrast image, objective detection in US images is relatively challenging. In this study, we employ the state-of-the-art RelTR~\cite{cong2023reltr} for ultrasound SG generation. This single-stage transformer-based approach eliminates the need for explicit object detection, enabling efficient and direct relationship extraction.

\par
To demonstrate the impact of SG representation in US imaging, this study introduces a novel method that leverages highly conceptualized SGs for realizing two critical tasks essential to advancing the democratization of portable and accessible US imaging: (1) generating graspable US explanation for ordinaries to self-learn anatomical and physiological knowledge, which is also particularly useful for POCUS scenarios, and (2) providing scanning guidance to reveal missing anatomies in the current imaging view, ensuring the displayed content aligns with user preferences. Both US summary~\cite{guo2024mmsummary,li2024ultrasound,zhou2025ultraadfinegrainedultrasoundanomaly} and probe guidance~\cite{zhao2022uspoint,men2022multimodal,xu2024transforming,jiang2024intelligent} tasks are important and have been investigated. However, to the best of our knowledge, this is the first work introducing SG and LLMs to boost intuitive US explanation and scanning guidance. 





\begin{figure}[t]
\centering
\includegraphics[width=\textwidth]{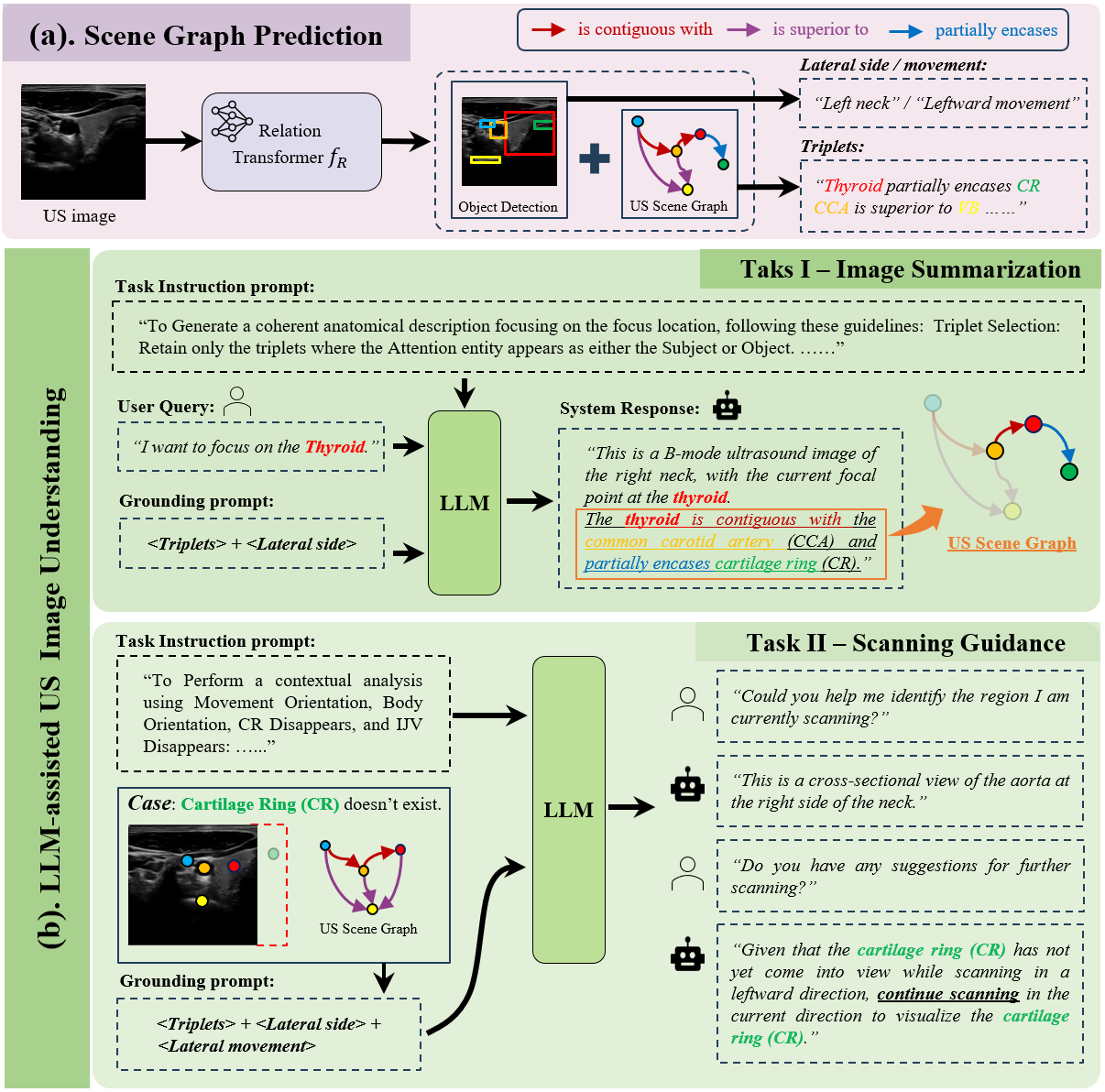}
\caption{An overview of the proposed framework. 
} 
\label{fig:overview}
\end{figure}

\section{Methodology}
In this section, we first introduce the process of SG prediction for US images acquired from the carotid artery scan. Then, we provide insights into how the predicted SG can be integrated into an LLM to facilitate US image understanding tasks, as illustrated in Fig.~\ref{fig:overview}.


\noindent
\textbf{Object Detection and Scene Graph Prediction} %
To predict a scene graph for a US image, triplets in the format of <$entity_1-predicate-entity_2$> should be defined to capture interactions between key anatomical structures within the scanning field of view. For cross-sectional carotid artery scanning, we select five representative anatomical structures as \textit{entities}: "Carotid Common Artery" (CCA), "Internal Jugular Vein" (IJV), "Cartilage Ring" (CR) above the Trachea, "Thyroid" (Th), and "Vertebral Body" (VB). Additionally, we define three interaction modes as \textit{predicates}: "is contiguous with", "partially encases", and "is superior to". These \textit{predicates} can effectively characterize the anatomical structure-based relationships among the five selected \textit{entities} in US images.


\par
Unlike conventional two-stage SG prediction methods, in this study, we employ the state-of-the-art RelTR~\cite{cong2023reltr}. RelTR follows a single-stage approach, which can simultaneously detect the \textit{entities} and predict SG rather than treating them as separate sequential steps. This design enhances efficiency by directly predicting relationships between anatomical structures without relying on intermediate procedures. As shown in Fig. \ref{fig:overview} (a), once an SG is predicted for the US image, it can subsequently be parsed into multiple triplet texts (referred to as \textit{triplets} for simplicity). Based on the object detection results and anatomical knowledge prior, we can identify whether the scan is performed on the neck's left or right \textit{lateral side}. Furthermore, by comparing two consecutive detections of the target anatomy, we can also identify the probe's \textit{lateral movement}. In the downstream tasks, the \textit{grounding prompt}, which enables the LLM to consider the current imaging results before responding to the user's query, will consist partially or entirely of the extracted \textit{triplets}, \textit{lateral side}, and \textit{lateral movement} information.


\noindent
\textbf{US Image Summarization}
This task aims to generate US summaries that emphasize the specified target of interest. In this context, given a user's query specifying a focus entity, a locally deployed LLM is tasked with providing a coherent summary of the US image. This summary includes a general description of the image, the focus area, and the relationships with adjacent \textit{entities}. These requirements are encapsulated into a fixed \textit{task instruction prompt}. To situate the LLM into the scanning loop, as shown in Fig. \ref{fig:overview} (b-Task I), a \textit{grounding prompt} comprising \textit{triplets} and \textit{lateral sides} is fed into the LLM to guide the task. This approach allows the LLM to understand the user's intent and implicitly prune the \textit{triplets}, retaining only the \textit{entities} directly related to the focus entity. By doing so, the LLM can generate coherent sentences that provide a personalized, intuitive explanation tailored to the region of the user's interest. This method enables even non-expert users to gain a clearer understanding of the US image and to learn anatomical knowledge of themselves. 



\noindent
\textbf{US Scanning Guidance}
Building upon the previous SG prediction and US image summarization tasks, the predicted SG can also be utilized to provide scanning guidance, helping non-expert users to operate a portable US probe to reveal the missing anatomies in the current imaging view during self-scanning. Similar to the US image summarization task, both the user queries about the desired anatomy to scan or the anatomy outside the imaging view, and a new \textit{task instruction prompt} for the US scanning guidance task will go through the LLM. However, to complete the task, in addition to the \textit{tirplets} and \textit{lateral sides}, the \textit{grounding prompt} also incorporates \textit{lateral movement} to indicate the relative motion direction of the US probe. The LLM then analyzes the scene graph to identify the missing \textit{entities} [see the case in Fig. \ref{fig:overview} (b-Task II) where CR is not present in the current US image]. By leveraging both the structural information from the SG and the scanning motion data indicated by \textit{lateral movement}, the LLM generates natural-language motion guidance, assisting the users in finding an imaging view that aligns with their preferences. 



\section{Experiments and Results Analysis}
\subsection{Implementation Details}
\noindent
\textbf{Model Selection}
We use hyperparameters similar to those in RelTR~\cite{cong2023reltr} for our experiments. The SG prediction network is trained for 800 epochs with a batch size of 16, on a workstation equipped with an RTX 4080 Super GPU. The initial learning rate for the transformer is \(10^{-4}\) and is reduced by a factor of 0.1 after 200 epochs to ensure stable convergence.
Since this work is aimed at potential applications in portable US devices, we prioritize ``lightweight" LLMs like LLaMa~\cite{touvron2023llama}, Qwen~\cite{bai2023qwen}, Gemma~\cite{team2024gemma}, Mathstral\footnote{https://mistral.ai/}
and the distilled DeepSeek R1 using the Qwen model (DS-R1-Qwen)~\cite{guo2025deepseek}. These lightweight models are quantized versions optimized for efficiency, reducing computational demands while maintaining performance, making them more suitable for real-time processing in resource-constrained environments. Additionally, we also employ high-capacity LLMs, such as
Gemini 2.0 Flash~\cite{team2023gemini}, and Grok 3 as reference models.



\noindent
\textbf{Data Acquisition}
The carotid artery US images were acquired using Siemens Juniper US Machine (ACUSON Juniper, SIEMENS AG, Germany) equipped with a 12L3 linear probe. The imaging and focus depths were set to 45 mm and 20 mm, respectively. A total of 289 US images were collected and annotated, with the training set comprising 262 images (resolution: 829$\times$770 pixels) from five volunteers. An additional 27 images collected from \textbf{different volunteers} were used for testing. Due to the lack of mature tools for scene graph annotation, we developed a lightweight annotation tool specifically for our 2D US dataset. It worth noting that, in addition to object detection annotations, scene graph annotation requires labeling triplets (<$subject-predicate-object$>), which makes the annotation process significantly more labor-intensive. Moreover, in SG prediction tasks, the \textit{predicate} in each triplet is closely tied to the spatial relationships between \textit{entities} in the image. As a result, standard data augmentation techniques used in traditional computer vision tasks are mostly inapplicable, with horizontal flipping being one of the few exceptions. These challenges have led to a relatively small dataset size.



\subsection{Object Detection and Scene Graph Prediction}
\subsubsection{Evaluation Metrics}
To evaluate object detection performance, we employ the widely adopted metric mean Average Precision (mAP). Specifically, we use two types of Average Precision (AP): AP@50, which is computed with an Intersection over Union (IoU) threshold of 50\%, and AP@[50:95], which averages precision across multiple IoU thresholds ranging from 50\% to 95\%, providing a stricter assessment, as outlined in the COCO evaluation protocol~\cite{lin2014microsoft}. For relation prediction, we employ Recall@K (R@K)~\cite{lu2016visual} and mean Recall@K (mR@K)~\cite{chen2019knowledge}. Given that the carotid US dataset contains a maximum of 7 relations, we set K = 5 and K = 20 for evaluation, respectively. 

\par
Given that the proposed method is intended for deployment on resource-constrained portable US devices, it is crucial to balance model size and predictive performance, ensuring the network remains compact while maintaining optimal accuracy. To achieve this, we conduct experiments to evaluate the effect of varying the number of encoder and decoder layers in RelTR’s transformer architecture~\cite{cong2023reltr} on both object detection and SG prediction. Table \ref{tab:sgg} summarizes the performance of object and relation detection across different configurations of transformer encoder and decoder layers. It is noted that the model with four layers achieved the best overall performance, yielding the highest AP@[50:95] (34.1\% vs. 32.4\% for the second best) and AP@50 (77.1\% vs. 70.3\%) in object detection. It also demonstrated consistently superior performance in relation detection, except for a marginal 0.4\% decrease in mR@5 compared to the five-layer model. While increasing the model size to five layers slightly enhanced relation detection in terms of mR@5, it resulted in a noticeable decline in both object and relation detection performance for all other evaluation metrics. Given the limited dataset, a four-layer encoder-decoder transformer strikes the best balance between object and relation detection, making it the most suitable choice for this setting. Therefore, we adopt this configuration as the default setting for all subsequent experiments.


\begin{table}[ht]
\centering
\caption{
Results for the object detection and scene graph detection 
}
\label{tab:sgg}
\resizebox{0.85\textwidth}{!}{%
\begin{tabular}{cccccccc}
\toprule
\multirow{2}{*}{No. of layers} &\multirow{2}{*}{Params}& \multicolumn{2}{c}{\textbf{Object Detection}} & \multicolumn{4}{c}{\textbf{Relation Detection}} \\

\cmidrule(lr){3-4} \cmidrule(lr){5-8}

 & & AP$_{50:95}\uparrow$ & AP$_{50}\uparrow$ & R@5$\uparrow$ & mR@5$\uparrow$ & R@20$\uparrow$ & mR@20$\uparrow$ \\
\midrule
3 layers &44M& 31.3 & 65.7 & 55.9 & 55.3 & 63.4 & 61.3 \\ 
4 layers &50M& \textbf{34.1} & \textbf{77.1} & \textbf{59.9} & 62.3 & \textbf{69.2} & \textbf{74.5} \\ 
5 layers &57M& 32.4 & 70.3 & 52.2 & \textbf{62.5} & 61.4 & 68.5 \\
\bottomrule
\end{tabular}
}
\end{table}

\subsection{Scene Graph-Powered US Understanding with LLMs} 
\textbf{Evaluation Metrics}
To evaluate the accuracy of LLM-generated text for the tasks of US image summarization and scanning guidance [see Fig. \ref{fig:overview} (b)], we use a combination of subjective assessment (referred to as $Acc$) and objective metrics. Subjective evaluation is conducted by third-party experts, who assess whether the LLM accurately follows the \textit{task instruction prompt} and executes the intended operation correctly. Objective evaluation, on the other hand, relies on widely used NLP metrics, including METEOR~\cite{banerjee2005meteor} and ROUGE$_{L}$~\cite{lin2004rouge}, which measure the linguistic similarity between the LLM-generated output and reference texts. The reference texts required for evaluation were generated using GPT-4o~\cite{achiam2023gpt}, followed by manual verification to ensure accuracy and reliability.

\begin{figure}[t]
\centering
\includegraphics[width=0.98\textwidth]{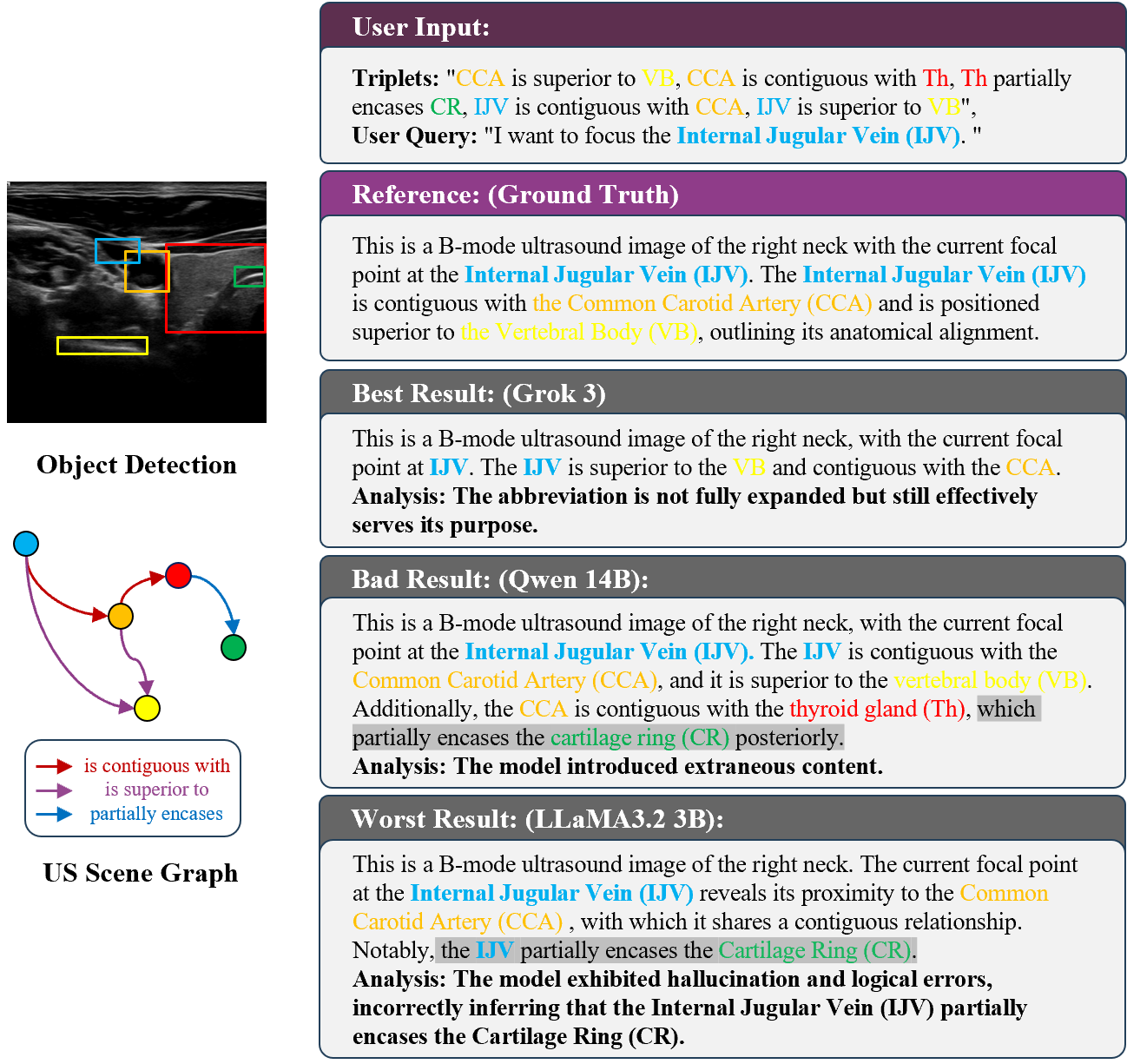}
\caption{An example of US image summarization generated by different LLMs in Task I, with key medical terms highlighted for clarity and comparison. Text highlighted with a gray background indicates areas containing significant errors. 
}
\label{fig:analysis}
\end{figure}

\noindent
\textbf{Task I: US Image Summarization}
The results (see Tab.~\ref{tab:downTask}) show that the large-scale models can significantly outperform lightweight quantized models in task completion, particularly Grok 3, which achieves the highest scores across all metrics. Among lightweight models, Qwen 2.5 (14B) and Gemma 2 (27B) achieve relatively strong performance but still fall behind large-capacity models by a noticeable margin in terms of the METEOR (Grok: 0.880 vs. 0.709 for the second best) and ROUGE$_L$ (Grok: 0.841 vs. 0.641 for the second best) metrics. These findings highlight a clear trend: as model size increases, improved reasoning capabilities enhance instruction execution, making large models the preferred choice for more complex and demanding tasks. For a more intuitive understanding of this trend, one can refer to the US image summarizations from different LLMs in Fig. \ref{fig:analysis} for further details.

\noindent
\textbf{Task II: Scanning Guidance}
Compared to Task I, Task II requires greater logical reasoning capabilities from the LLMs to ensure accurate execution. As a result, all models exhibit lower overall scores across Accuracy, METEOR, and ROUGE$_{L}$. Among all models, Grok 3 continues to outperform the others ($Acc$: 0.776 vs. 0.633 of the second best), showcasing its superior ability to handle complex reasoning tasks. However, lightweight quantized models such as Qwen 2.5 (14B) and Gemma 2 (27B) maintain relatively high accuracy, similar to their performance in Task I. Therefore,
in resource-constrained settings, Qwen 2.5 (14B) and Gemma 2 (27B) can offer a practical solution; with appropriate quantization for local deployment, these two models can perform inference using only 8GB and 14GB of VRAM, respectively.




\begin{table}[t]
\centering
\caption{Evaluation results of different LLM models on Task I and Task II, using Accuracy ($Acc$), METEOR, and ROUGE$_{L}$. The parentheses indicate model parameter sizes (e.g., LLaMA 3.2 (1B) = 1B parameters). $\dagger$: High-capacity LLMs.}
\label{tab:downTask}
\resizebox{0.90\textwidth}{!}{%
\begin{tabular}{lccccccc}
\toprule
\multicolumn{1}{c}{\multirow{2}{*}{Model}} & \multicolumn{3}{c}{\textbf{Task I}} & \multicolumn{3}{c}{\textbf{Task II}}\\

\cmidrule(lr){2-4} \cmidrule(lr){5-7} 

 &~~Acc~$\uparrow$ & METEOR$\uparrow$ & ROUGE$_{L}$$\uparrow$ &~~Acc~$\uparrow$ &METEOR$\uparrow$&ROUGE$_{L}$$\uparrow$  \\
\midrule
LLaMA 3.2(1B) &0.265 & 0.550 & 0.387& 0.408& 0.392& 0.300 \\ %
LLaMA 3.2(3B)  &0.531 & 0.534 & 0.365& 0.347& 0.403& 0.390 \\ %
LLaMA 3.1(8B)  &0.735 & 0.489 & 0.335& \textbf{0.633}& 0.395& 0.290 \\ %
Mathstral v0.1(7B) &0.612 & 0.545 & 0.384& 0.327& \textbf{0.447}& \textbf{0.496} \\ %
DS-R1-Qwen(7B) &0.551 & 0.590 & 0.576& 0.265& 0.400& 0.478 \\
Qwen 2.5(14B) &0.755 & \textbf{0.709} & \textbf{0.641}& 0.429& 0.404& 0.423 \\
Gemma 2(27B) &\textbf{1.000} & 0.590 & 0.615& 0.469& 0.401& 0.508 \\
\midrule
$^\dagger$Gemini 2.0 Flash &0.980 & 0.589 & 0.736& 0.592& 0.452& 0.623 \\
$^\dagger$Grok 3 &\textbf{1.000}& \textbf{0.880} & \textbf{0.841}& \textbf{0.776}& \textbf{0.490}& \textbf{0.665} \\
\bottomrule
\end{tabular}
}

\end{table}


\section{Discussion and Conclusion}
This study introduces semantic scene graphs for ultrasound images to provide intuitive image explanations and effective scanning guidance for individuals with limited physiological knowledge. The transformer-based RelTR model is used to predict the semantic SG of US images, capturing key anatomical structures and their relationships. This information-rich SG is further used with recent advanced LLMs to demonstrate that the system can help non-expert users better understand and analyze US images. While the proposed framework shows promising results for carotid artery scans, the current method is only validated on carotid images. To have a robust performance across different anatomies, a large dataset, including images of different anatomies, should be collected.  

\par
Nevertheless, this paper demonstrates the promising potential of an SG-guided LLM framework for interpreting US images and providing scanning guidance. Given the inherent scarcity of ultrasound data, the proposed framework offers a practical and innovative alternative to large-scale vision-language models (VLMs) that demand extensive ultrasound datasets for training. Furthermore, these applications show great potential for promoting self-learning of anatomical and physiological knowledge, especially among young individuals.

\begin{credits}

\end{credits}

\clearpage

%
%
%
\newpage

\bibliographystyle{splncs04}
\bibliography{reference}

\begin{thebibliography}{10}
\providecommand{\url}[1]{\texttt{#1}}
\providecommand{\urlprefix}{URL }
\providecommand{\doi}[1]{https://doi.org/#1}

\bibitem{achiam2023gpt}
Achiam, J., Adler, S., Agarwal, S., Ahmad, L., Akkaya, I., Aleman, F.L., Almeida, D., Altenschmidt, J., Altman, S., Anadkat, S., et~al.: Gpt-4 technical report. arXiv preprint arXiv:2303.08774  (2023)

\bibitem{andersen2019point}
Andersen, C.A., Holden, S., Vela, J., Rathleff, M.S., Jensen, M.B.: Point-of-care ultrasound in general practice: a systematic review. The Annals of Family Medicine  \textbf{17}(1),  61--69 (2019)

\bibitem{bai2023qwen}
Bai, J., Bai, S., Chu, Y., Cui, Z., Dang, K., Deng, X., Fan, Y., Ge, W., Han, Y., Huang, F., et~al.: Qwen technical report. arXiv preprint arXiv:2309.16609  (2023)

\bibitem{banerjee2005meteor}
Banerjee, S., Lavie, A.: Meteor: An automatic metric for mt evaluation with improved correlation with human judgments. In: Proceedings of the acl workshop on intrinsic and extrinsic evaluation measures for machine translation and/or summarization. pp. 65--72 (2005)

\bibitem{bi2024machine}
Bi, Y., Jiang, Z., Duelmer, F., Huang, D., Navab, N.: Machine learning in robotic ultrasound imaging: Challenges and perspectives. Annual Review of Control, Robotics, and Autonomous Systems  \textbf{7} (2024)

\bibitem{chen2019knowledge}
Chen, T., Yu, W., Chen, R., Lin, L.: Knowledge-embedded routing network for scene graph generation. In: CVPR. pp. 6163--6171 (2019)

\bibitem{cong2023reltr}
Cong, Y., Yang, M.Y., Rosenhahn, B.: Reltr: Relation transformer for scene graph generation. IEEE Transactions on Pattern Analysis and Machine Intelligence  \textbf{45}(9),  11169--11183 (2023)

\bibitem{dhamo2020semantic}
Dhamo, H., Farshad, A., Laina, I., Navab, N., Hager, G.D., Tombari, F., Rupprecht, C.: Semantic image manipulation using scene graphs. In: CVPR. pp. 5213--5222 (2020)

\bibitem{guo2025deepseek}
Guo, D., Yang, D., Zhang, H., Song, J., Zhang, R., Xu, R., Zhu, Q., Ma, S., Wang, P., Bi, X., et~al.: Deepseek-r1: Incentivizing reasoning capability in llms via reinforcement learning. arXiv preprint arXiv:2501.12948  (2025)

\bibitem{guo2024mmsummary}
Guo, X., Men, Q., Noble, J.A.: Mmsummary: Multimodal summary generation for fetal ultrasound video. In: MICCAI. pp. 678--688. Springer (2024)

\bibitem{huh2023breast}
Huh, J., Park, H.J., Ye, J.C.: Breast ultrasound report generation using langchain. arXiv preprint arXiv:2312.03013  (2023)

\bibitem{im2024egtr}
Im, J., Nam, J., Park, N., Lee, H., Park, S.: Egtr: Extracting graph from transformer for scene graph generation. In: CVPR. pp. 24229--24238 (2024)

\bibitem{islam2020learning}
Islam, M., Seenivasan, L., Ming, L.C., Ren, H.: Learning and reasoning with the graph structure representation in robotic surgery. In: MICCAI. pp. 627--636. Springer (2020)

\bibitem{yang2018graph}
J, Y., Lu, J., Lee, S., Batra, D., Parikh, D.: Graph r-cnn for scene graph generation. In: ECCV. pp. 670--685 (2018)

\bibitem{jiang2024intelligent}
Jiang, Z., Bi, Y., Zhou, M., Hu, Y., Burke, M., Navab, N.: Intelligent robotic sonographer: Mutual information-based disentangled reward learning from few demonstrations. The International Journal of Robotics Research  \textbf{43}(7),  981--1002 (2024)

\bibitem{jiang2023robotic}
Jiang, Z., Salcudean, S.E., Navab, N.: Robotic ultrasound imaging: State-of-the-art and future perspectives. Medical image analysis  \textbf{89},  102878 (2023)

\bibitem{li2024ultrasound}
Li, J., Su, T., Zhao, B., Lv, F., Wang, Q., Navab, N., Hu, Y., Jiang, Z.: Ultrasound report generation with cross-modality feature alignment via unsupervised guidance. IEEE Transactions on Medical Imaging  (2024)

\bibitem{lin2004rouge}
Lin, C.Y.: Rouge: A package for automatic evaluation of summaries. In: Text summarization branches out. pp. 74--81 (2004)

\bibitem{lin2014microsoft}
Lin, T., Maire, M., Belongie, S., Hays, J., Perona, P., Ramanan, D., Doll{\'a}r, P., Zitnick, C.L.: Microsoft coco: Common objects in context. In: Computer vision--ECCV 2014: 13th European conference, zurich, Switzerland, September 6-12, 2014, proceedings, part v 13. pp. 740--755. Springer (2014)

\bibitem{lu2016visual}
Lu, C., Krishna, R., Bernstein, M., Fei-Fei, L.: Visual relationship detection with language priors. In: Computer Vision--ECCV 2016: 14th European Conference, Amsterdam, The Netherlands, October 11--14, 2016, Proceedings, Part I 14. pp. 852--869. Springer (2016)

\bibitem{maier2017surgical}
Maier-Hein, L., Vedula, S.S., Speidel, S., Navab, N., Kikinis, R., Park, A., Eisenmann, M., Feussner, H., Forestier, G., Giannarou, S., et~al.: Surgical data science for next-generation interventions. Nature Biomedical Engineering  \textbf{1}(9),  691--696 (2017)

\bibitem{men2022multimodal}
Men, Q., Teng, C., Drukker, L., Papageorghiou, A.T., Noble, J.A.: Multimodal-guidenet: Gaze-probe bidirectional guidance in obstetric ultrasound scanning. In: International Conference on Medical Image Computing and Computer-Assisted Intervention. pp. 94--103. Springer (2022)

\bibitem{murali2023latent}
Murali, A., Alapatt, D., Mascagni, P., Vardazaryan, A., Garcia, A., Okamoto, N., Mutter, D., Padoy, N.: Latent graph representations for critical view of safety assessment. IEEE Transactions on Medical Imaging  \textbf{43}(3),  1247--1258 (2023)

\bibitem{ockenden2024role}
Ockenden, E.S., Frischer, S.R., Cheng, H., Noble, J.A., Chami, G.F.: The role of point-of-care ultrasound in the assessment of schistosomiasis-induced liver fibrosis: A systematic scoping review. PLOS Neglected Tropical Diseases  \textbf{18}(3),  e0012033 (2024)

\bibitem{ozsoy20224d}
{\"O}zsoy, E., {\"O}rnek, E.P., Eck, U., Czempiel, T., Tombari, F., Navab, N.: 4d-or: Semantic scene graphs for or domain modeling. In: MICCAI. pp. 475--485. Springer (2022)

\bibitem{sanner2024voxel}
Sanner, A.P., Grauhan, N.F., Brockmann, M.A., Othman, A.E., Mukhopadhyay, A.: Voxel scene graph for intracranial hemorrhage. In: MICCAI. pp. 519--529. Springer (2024)

\bibitem{team2023gemini}
Team, G., Anil, R., Borgeaud, S., Alayrac, J.B., Yu, J., Soricut, R., Schalkwyk, J., Dai, A.M., Hauth, A., Millican, K., et~al.: Gemini: a family of highly capable multimodal models. arXiv preprint arXiv:2312.11805  (2023)

\bibitem{team2024gemma}
Team, G., Riviere, M., Pathak, S., Sessa, P.G., Hardin, C., Bhupatiraju, S., Hussenot, L., Mesnard, T., Shahriari, B., Ram{\'e}, A., et~al.: Gemma 2: Improving open language models at a practical size. arXiv preprint arXiv:2408.00118  (2024)

\bibitem{touvron2023llama}
Touvron, H., Lavril, T., Izacard, G., Martinet, X., Lachaux, M.A., Lacroix, T., Rozi{\`e}re, B., Goyal, N., Hambro, E., Azhar, F., et~al.: Llama: Open and efficient foundation language models. arXiv preprint arXiv:2302.13971  (2023)

\bibitem{xu2024transforming}
Xu, H., Wu, J., Cao, G., Chen, Z., Lei, Z., Liu, H.: Transforming surgical interventions with embodied intelligence for ultrasound robotics. In: MICCAI. pp. 703--713. Springer (2024)

\bibitem{yuan2024advancing}
Yuan, K., Kattel, M., Lavanchy, J.L., Navab, N., Srivastav, V., Padoy, N.: Advancing surgical vqa with scene graph knowledge. International Journal of Computer Assisted Radiology and Surgery  \textbf{19}(7),  1409--1417 (2024)

\bibitem{zhao2022uspoint}
Zhao, C., Droste, R., Drukker, L., Papageorghiou, A.T., Noble, J.A.: Uspoint: Self-supervised interest point detection and description for ultrasound-probe motion estimation during fine-adjustment standard fetal plane finding. In: MICCAI. pp. 104--114. Springer (2022)

\bibitem{zhou2025ultraadfinegrainedultrasoundanomaly}
Zhou, Y., Bi, Y., Tong, W., Wang, W., Navab, N., Jiang, Z.: Ultraad: Fine-grained ultrasound anomaly classification via few-shot clip adaptation (2025), \url{https://arxiv.org/abs/2506.19694}

\end{thebibliography}
%




\end{document}